\documentclass{article}

\usepackage[preprint]{neurips_2025}

\usepackage{enumitem}
\usepackage[utf8]{inputenc} %
\usepackage[T1]{fontenc}    %
\usepackage{hyperref}       %
\usepackage{url}            %
\usepackage{booktabs}       %
\usepackage{amsfonts}       %
\usepackage{amssymb}
\usepackage{nicefrac}       %
\usepackage{microtype}      %
\usepackage{xcolor}         %
\usepackage{xspace}
\usepackage{graphicx}
\usepackage{wrapfig}
\usepackage{subfigure}
\usepackage{amsmath}
\usepackage{multirow}
\usepackage{listings}
\usepackage{makecell}
\usepackage{tcolorbox}
\tcbset{
  colback=gray!10,    %
  colframe=black,     %
  boxsep=4pt,         %
  left=6pt,           %
  right=6pt,          %
}

\newcommand{\ie}{\emph{i.e.,}\xspace}

\newcommand{\eg}{\emph{e.g.,}\xspace}

\newcommand{\ignore}[1]{}

\newcommand{\model}{Vision-G1\xspace}
\newcommand{\red}[1]{\textcolor{red}{#1}}

\title{\model: Towards General Vision Language Reasoning with Multi-Domain Data Curation}

\author{%
  Yuheng Zha\thanks{Equal contribution.}$\ \ ^\spadesuit$\quad Kun Zhou\thanks{Corresponding author: kuzhou@ucsd.edu}\xspace\xspace$^{*}$$^{\spadesuit}$ \quad Yujia Wu $^\spadesuit$ \quad Yushu Wang$^\spadesuit$ \quad Jie Feng$^\spadesuit$ \\ \textbf{Zhi Xu$^\spadesuit$ \quad Shibo Hao$^\spadesuit$ \quad Zhengzhong Liu$^\diamondsuit$ \quad Eric P. Xing$^{\clubsuit \diamondsuit}$ \quad Zhiting Hu$^\spadesuit$} \\
  UC San Diego$^\spadesuit$, Carnegie Mellon University$^\clubsuit$, MBZUAI$^\diamondsuit$
  \\
}

\begin{document}

\maketitle

\begin{abstract}

Recent vision-language models (VLMs) show strong reasoning capabilities through training with reinforcement learning from verifiable rewards (RLVR). 
 Despite their success, current training pipelines for reasoning VLMs focus on a limited range of tasks, such as mathematical and logical reasoning. As a result, these models face difficulties in generalizing their reasoning capabilities to a wide range of domains, primarily due to the scarcity of readily available and verifiable reward data beyond these narrowly defined areas. Moreover, integrating data from multiple domains is challenging, as the compatibility between domain-specific datasets remains uncertain.
To address these limitations, we build a comprehensive RL-ready visual reasoning dataset from 46 data sources across 8 dimensions, covering a wide range of tasks such as infographic, mathematical, spatial, cross-image, graphic user interface, medical, common sense and general science. 
We propose an influence function based data selection and difficulty based filtering strategy to identify high-quality training samples from this dataset. 
Subsequently, we train the VLM, referred to as \model, using multi-round RL with a data curriculum to iteratively improve its visual reasoning capabilities.
Our model achieves state-of-the-art performance across various visual reasoning benchmarks, outperforming similar-sized VLMs and even proprietary models like GPT-4o and Gemini-1.5 Flash. The model, code and dataset are publicly available at \url{https://github.com/yuh-zha/Vision-G1}.

\end{abstract}

\section{Introduction}
Large language models (LLMs) trained with reinforcement learning (RL) from verifiable rewards, such as DeepSeek R1 \citep{guo2025deepseekr1}, show strong reasoning capabilities on diverse tasks such as math \cite{cobbe2021gsm8k,hendrycksmath2021} and coding \citep{jimenez2024swebench}. Following this paradigm, the open source community has proposed additional reasoning LLM training methods \citep{yu2025dapo,hu2025openreasoner, yuan2025vapo, guru} to advance these capabilities further. It is promising to apply similar methods from pure language models to vision language models (VLMs), enabling VLMs to exhibit strong reasoning capabilities on a wide range of visual reasoning tasks. While the common practice for training vision-language models \citep{bai2025qwen25vl, chen2024internvl, llava} involves only supervised fine-tuning after pre-training,
there hav been some initial attempts to post-train VLMs with reinforcement learning from verifiable rewards to enhance their visual perception \citep{liu2025visualrft,shen2025vlmr1} and reasoning \citep{peng2025lmmr1, meng2025mmeureka,du2025virgo,yang2025r1onevision,zhan2025visionr1,huang2025visionr1,wang2025thinklite,wang2025vlrethinker,team2025kimi} capabilities. For example, by collecting K12-level exam questions with verifiable answers, MM-Eureka \citep{meng2025mmeureka} trains a VLM to improve its math and science-related reasoning capabilities.

Despite their success, VLMs continue to struggle with visual reasoning tasks in broader domains, which require multiple aspects of reasoning, including logical, commonsense, and physical knowledge \citep{zhang2024vlmcot,xu2025llavacot,chen2024spatialvlm}. 
Such issue stems from the restricted data types that the model can interact with during the training.
Though several works \citep{wang2025thinklite, wang2025vlrethinker, xu2025mixed, liang2025modomodo} have tried collecting data from domains beyond mathematical reasoning and converting them into verifiable formats for training, the optimal data-mixing paradigm for comprehensively enhancing VLM reasoning capabilities remains unclear.

In this work, we propose a pipeline for training a vision language model using reinforcement learning with verifiable rewards, aimed at enhancing its reasoning capabilities across general domains.
To generalize its reasoning capabilities to broader domains, we build a large RL-ready training dataset covering 8 domains: infographic reasoning, graphic user interface (web), mathematical reasoning, cross-image reasoning, spatial reasoning, medical, general science, and common sense. It consists of 46 visual reasoning datasets and 13 sub-tasks in total. For each data source, we 
filter out instances with non-verifiable answers (\eg open-ended questions), retaining only those with numeric values, multiple-choice options, yes/no answers, or other single-word ground truths. 

\begin{wrapfigure}{r}{0.6\textwidth}
  \centering
  \includegraphics[width=\linewidth]{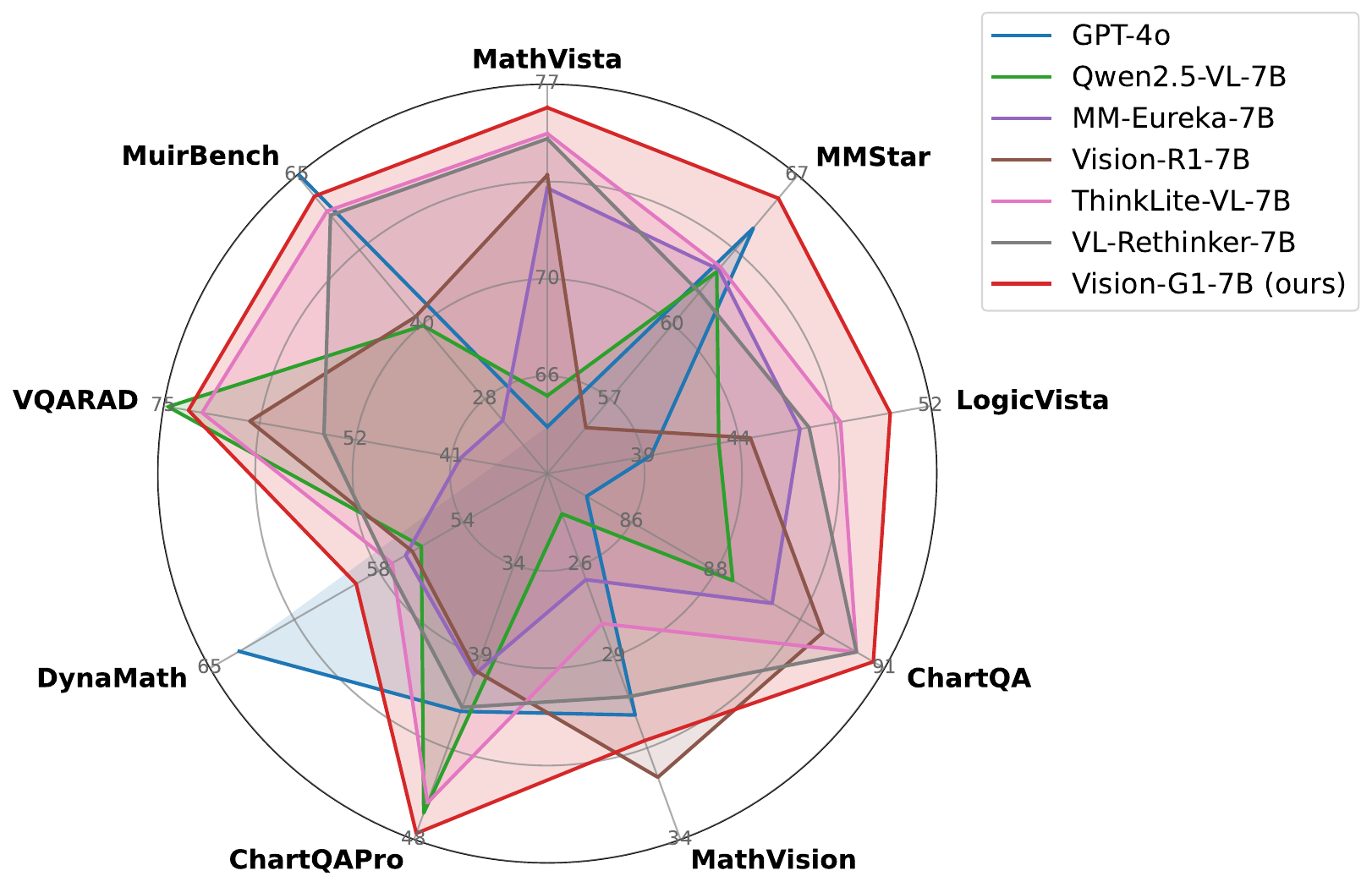}
  \caption{Radar chart showing the performance of our proposed \model and other vision language models on ten visual reasoning benchmarks. The baselines include RL-trained models and GPT-4o.}
  \label{fig:myfig}
\end{wrapfigure}

The collected raw datasets generally contain instances of varying quality and quantity. Simply mixing them admits low-quality instances (\eg too easy or too hard), impeding effective learning of general visual reasoning knowledge.
Meanwhile, existing approaches \citep{shen2025vlmr1,peng2025lmmr1} largely rely on heuristic strategies or human-crafted features for filtering, requiring specific manual designs and limiting adaptability to the above heterogeneous datasets.
To address this issue, we propose a data filtering method based on the influence function~\cite{xia2024less,Pruthi2020EstimatingTD}, aimed at removing unhelpful instances from the RL training data.
Concretely, we use reject sampling on a small subset of the training data to obtain high-quality responses from the initial VLM. We then apply the influence function to the remaining training set to remove instances with negative influence values.

To further improve the reasoning capability obtained from the filtered multi-source datasets, we develop a data curriculum strategy for multi-round RL training.
Specifically, we sample rollouts from the VLM in the previous training round and estimate the instance-level difficulty by computing the average rollout accuracy. Moderately difficult instances matching the current VLM’s capabilities are then selected for the next round of training.
We iteratively apply the data selection and RL training to progressively improve our \model's general reasoning capability.
Our \model-7B achieves state-of-the-art performance among competitive baselines on 17 benchmarks, spanning comprehensive visual reasoning, math-related reasoning, and domain-specific reasoning tasks. 
Ablation studies show that our method effectively selects high-quality data and improves the training of the general reasoning vision language model.

\section{Related Work}
We briefly introduce related work from the following three aspects: the background of VLMs, reasoning in VLMs, and data selection methods for language model training.

\paragraph{Vision-Language Models.}
Large language models continue to advance, with GPT-4 \citep{openai2024gpt4o} and Qwen-2.5-VL \citep{bai2025qwen25vl} exhibiting emergent skills such as in-context learning and sophisticated reasoning. Building on this progress, vision-language models (VLMs) extend these abilities to multimodal inputs by coupling a vision backbone (\eg a Vision Transformer) with an LLM-based text decoder, allowing unified reasoning over images and text. 
Recent work has further strengthened both perception and reasoning.  LLaVA-NeXT, for example, supports variable-resolution input by tiling images into adaptive grids \citep{llava}, whereas Qwen2-VL introduces M-RoPE, a refined rotary position encoding that unifies spatial and temporal cues for images and video \citep{bai2025qwen25vl}.  Several systems treat pictures and clips within a single architecture and merge their instruction data during fine-tuning \citep{xu2024llavacot,zhu2025internvl3}.  On the reasoning front, models such as QvQ \citep{qvq} and Virgo \citep{du2025virgo} push performance on complex tasks by generating extended chains of thought.

\paragraph{Reasoning Vision-Language Models.}
Building on breakthroughs in large reasoning language models such as OpenAI o1~\citep{OpenAIo1} and DeepSeek-R1~\citep{guo2025deepseekr1}, recent work has turned to strengthening the reasoning capabilities of Vision-Language Models (VLMs).
Early approaches~\citep{xu2024llavacot,du2025virgo} assemble multimodal Chain-of-Thought (CoT) datasets and employ supervised fine-tuning to boost the reasoning ability of VLMs.
Motivated by the success of reinforcement learning techniques~\citep{shao2024deepseekmath,deepseekai2025r1}, recent studies have used RL with task-specific, verifiable reward schemes (\eg answer accuracy and detection IoU) to provide supervisory signals~\citep{chen2025r1v,liu2025visualrft}, improving VLM reasoning and exhibiting remarkable performance.
However, existing work has found that relying solely on RL often fails to elicit the long chain-of-thought reasoning ability of VLMs.
To address this, supervised fine-tuning~\citep{meng2025mmeureka,chen2025vlaathinker} and special prompting mechanisms~\citep{wei2023cot,zhang2024vlmcot} are proposed to encourage the long CoT generation style.

\paragraph{Data Selection for Language Model Training.}
To train large language models~(LLMs) and vision-language models~(VLMs), choosing the right data is always critical.
Existing data selection methods~\citep{zhou2024lima,xia2024less} focus on removing the redundant or harmful instances to reduce the training cost and improve the stability.
Early work~\citep{zhou2024lima,chen2023maybe} mostly relies on human experience to design heuristic rules, and shows that a high-quality training small dataset is able to learn specific capabilities, \eg instruction following and human alignment.
Subsequent methods leverage the features that can be computed by simple metrics or LLMs (\eg length, complexity, and diversity), for data value estimation and selection~\citep{jain2023llm,liu2023makes,zhuo2024astraios,muennighoff2023octopack}.
However, the above features need specific designs for diverse tasks, making them hard to handle a highly heterogeneous mix of multi-task datasets.
To solve it, influence function methods~\citep{Pruthi2020EstimatingTD} have been proposed, which can estimate the influence of each training instance on other ones.
Recent work has simplified the influence estimation function into a simple gradient similarity computation formula, and exhibited remarkable performance on text and visual instruction selection~\citep{xia2024less,zhou2024lima}.
In this work, we utilize the influence function for filtering low-quality instances. Besides, we also use the rollout accuracy to estimate the difficulty for immediate high-value training data selection in the multi-round RL training process.

\begin{figure}[t]
    \centering
    \includegraphics[width=1\textwidth]{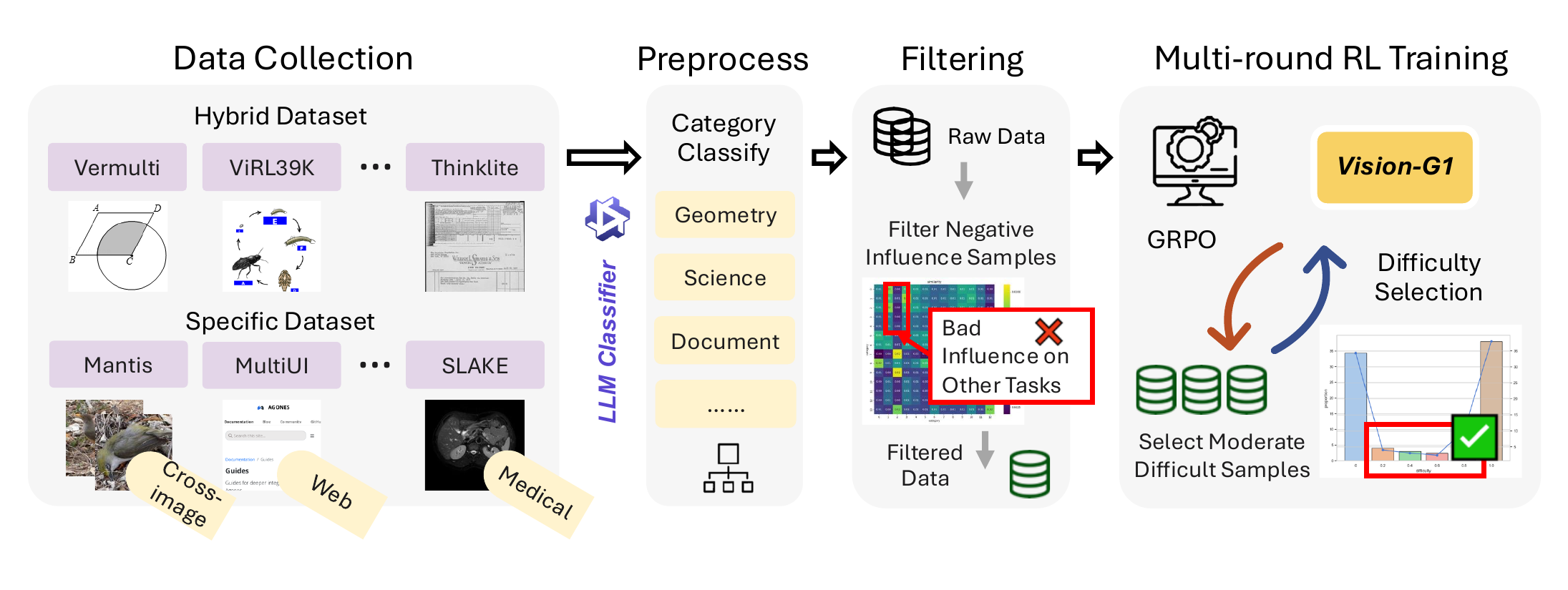} 
    \caption{The overview of our approach, consisting of collecting and preprocessing a mixture of heterogeneous datasets, low-quality instances filtering based on influence function, and multi-round RL training with difficulty-based data selection strategy.}
    \label{fig:approach}
\end{figure}
\section{Method}
In this section, we present the methodology employed to train the \model model. Specifically, we leverage the influence function and difficulty-based filtering to curate high-quality samples from multi-modal reasoning datasets spanning eight domains (Section \ref{sec:data_build}). Subsequently, we perform multi-round reinforcement learning (RL) on the base model to enhance its reasoning capabilities (Section \ref{sec:multi_round_training}). An overview of the data processing and training pipeline is illustrated in Figure \ref{fig:approach}.

\subsection{Training Corpus Construction} \label{sec:data_build}
\begin{wrapfigure}[25]{r}{0.5\textwidth}
  \centering
  \includegraphics[width=\linewidth]{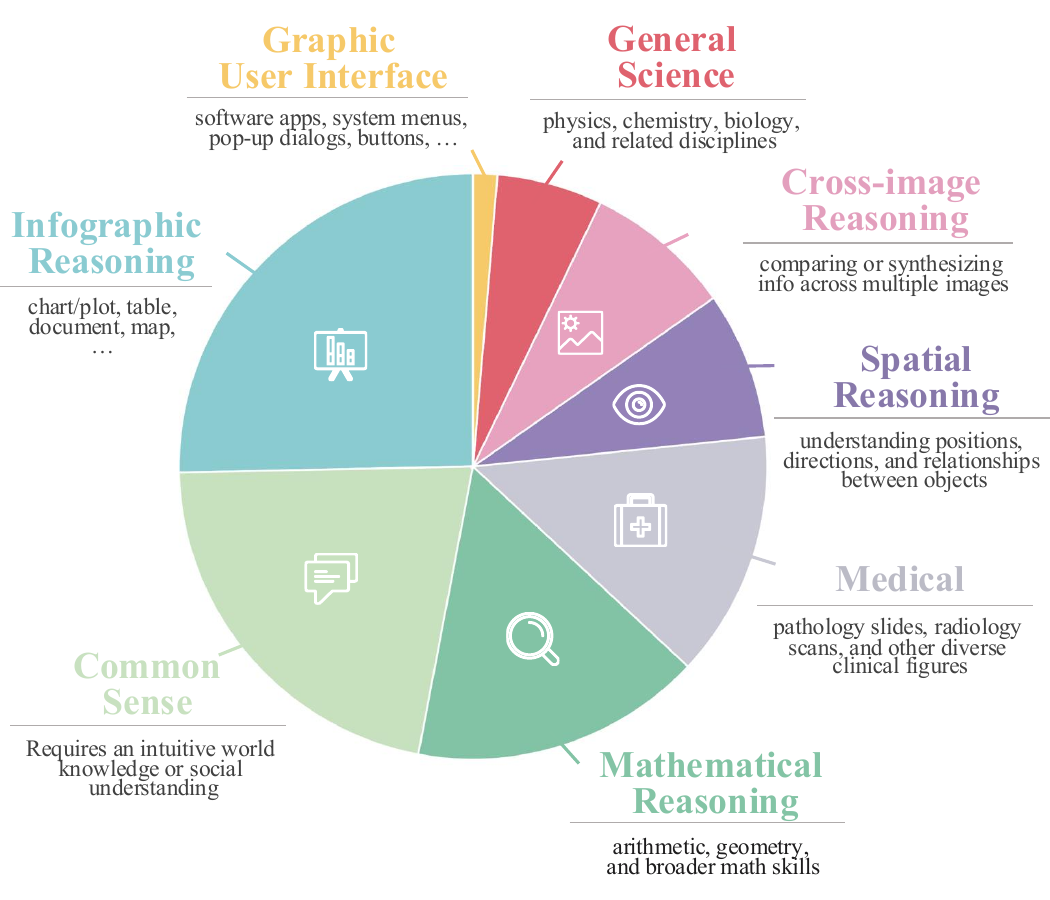}
  \caption{Source dataset distribution of our \model. KR: knowledge reasoning, IR: infographic reasoning, MR: mathematical reasoning, CIR: cross-image reasoning, SR: spatial reasoning. The full list of source datasets is shown in Table \ref{tab:data_sources}.}
  \label{fig:taxonomy}
\end{wrapfigure}

To construct a comprehensive dataset for model training, we compile a diverse set of visual reasoning datasets spanning various domains and tasks, and preprocess their instances into a unified format with category labels. Figure \ref{fig:taxonomy} shows the distribution of the source datasets.

\paragraph{Data Domains.} To enhance the model’s capabilities beyond mathematical reasoning, we incorporate datasets from the following domains: infographic reasoning, graphic user interfaces (web), mathematical reasoning, cross-image reasoning, spatial reasoning, general science, common sense, and medicine. \textit{\textbf{Infographic reasoning}} encompasses tasks involving charts, documents, and maps. \textit{\textbf{Graphic user interface}} tasks focus specifically on web pages, where text, graphics, and images appear in an interleaved manner. \textit{\textbf{Mathematical reasoning}} covers problems such as geometry and arithmetic. \textit{\textbf{Cross-image reasoning}} refers to tasks that require understanding relationships across multiple images. \textit{\textbf{Spatial reasoning}} involves interpreting the spatial relationships between objects within an image. \textit{\textbf{General science}} includes problems from disciplines such as physics and chemistry. \textit{\textbf{Common Science}} is a category for everyday questions such as understanding traffic signals or laundry care symbols. \textit{\textbf{Medical}} tasks primarily involve reasoning over X-ray and pathology images. For the full list of collected datasets, please refer to Table \ref{tab:data_sources}.

\paragraph{Data Preprocess.}
Data from different sources often appear in heterogeneous formats, while the training process specifically requires images and corresponding questions as model inputs, along with ground-truth annotations that serve as reward signals. To address this, we convert all data items into a standardized format. Furthermore, to characterize the domain distribution of the collected data, we assign category labels to each data instance.

$\bullet$ \emph{Format Unification.}
We extract questions, images, and ground truth from each dataset instance. The ground truths are checked with a rule-based function to ensure they are verifiable with rule-based reward models. Unverifiable instances are filtered out. We concatenate the questions with a prompt that promotes thinking, following Thinklite \citep{wang2025thinklite}. We also instruct the model to output the answer in a structured format, \ie in \textbackslash boxed\{\}. The full prompt and the details are shown in Appendix \ref{app:format}.

$\bullet$ \emph{Category Classification.}
To effectively track data proportions and knowledge distribution, we establish a category taxonomy and classify all instances within the collected datasets.
We identify 13 fine-grained dimensions within those eight task domains, and obtain a hierarchical category taxonomy (shown in Fig.~\ref{fig:taxonomy}) to facilitate classification.
Finally, we use a VLM (\ie Qwen2.5VL-32B-Instruct) classifier to category all data instances into the above dimensions. The classifier prompt is shown in Table \ref{tab:classifier_prompt}.

\subsection{Multi-round Reinforcement Learning with Data Curriculum} \label{sec:multi_round_training}
We train our \model using multi-round reinforcement learning with a data curriculum. Specifically, we begin by removing low-quality data through influence-function analysis and difficulty-based filtering. To progressively enhance \model’s reasoning capabilities, we conduct multi-round RL training interleaved with our data selection strategy.

\paragraph{Influence Function based Data Selection.}
We first use the influence function to estimate the impact between different data domains. The influence \texttt{I} of an instance $z$ on another one $z'$, for a model parameterized by $\theta$, can be estimated by computing the similarity between their gradients \citep{Pruthi2020EstimatingTD}, denoted as:
\begin{equation}
    \texttt{I}(z,z')\propto \textit{Sim}(\nabla l(z, \theta), \nabla l(z', \theta)), \label{eq-influ}
\end{equation}
where $l(\cdot, \cdot)$ denotes the cross-entropy loss function. 
Concretely, we formulate the RL influence estimation function by modeling the influence between instances in the training dataset. For an instance $z$ in the dataset:
\begin{equation}
I(z)
= \frac{1}{\bigl|\mathcal{D}_{\operatorname{dom}(z)}\bigr|}
  \sum_{z' \in \mathcal{D}_{\operatorname{dom}(z)}} I(z,z')
\;+\;
  \frac{1}{\bigl|\mathcal{D}\setminus \mathcal{D}_{\operatorname{dom}(z)}\bigr|}
  \sum_{z' \in \mathcal{D}\setminus \mathcal{D}_{\operatorname{dom}(z)}} I(z,z'). \label{eq:infl-data}
\end{equation}
where $\mathcal{D}$ is the full dataset, and $\operatorname{dom}(z)$ is the domain label of $z$ (\eg infographic reasoning). $\mathcal{D}_{\operatorname{dom}(z)}$ is defined as:
$
\mathcal{D}_{\operatorname{dom}(z)}
:= \{\, z' \in \mathcal{D} \mid \operatorname{dom}(z') = \operatorname{dom}(z) \,\}.
$ 
To reduce the cost for computing the gradient, we fine-tune a LoRA \citep{hu2022lora} module on high-quality reasoning chains, obtained via reject sampling from the base VLM on a subset of the training data.
For estimating the instance influence in the training set, we first sample the rollouts from the base VLM and compute the gradients on LoRA parameters using Equation \ref{eq:infl-data}.
Next, we perform random projection to obtain the low-dimensional features following \cite{xia2024less}, and we use cosine similarity to estimate the influence of each instance (Eq.~\ref{eq-influ}).
Finally, we filter the instances with low influence and ensure that the remaining instances of each dimension are uniformly distributed.

\begin{table}[t]
\centering
\small
\caption{Benchmarking results for general visual reasoning tasks. The best and second-best ones among 7B VLMs are marked in \textbf{bold} and \underline{underlined}, respectively.}

\resizebox{\textwidth}{!}{
\begin{tabular}{lccccccc}
\toprule
\textbf{Models}  & \textbf{MathVista}& \textbf{MMMU-Val} & \textbf{MMMU-Pro} & \textbf{MMStar} & \textbf{LogicVista} &\textbf{ChartQA} \\ 
\midrule
\multicolumn{7}{c}{\textit{Proprietary Vision-Language Models}} \\
\midrule
GPT-4o &  63.8& 69.1 &51.9  &64.7 & 39.6 &85.7  \\
Claude-3.5 & 67.7&68.3&51.5& 65.1& 44.4&90.8\\
Gemini-1.5 Flash & 58.4 & 56.1& - & - &40.0 & 79.0\\ %
Gemini-1.5 Pro & 63.9 & 65.8 & 46.9 &59.1 & 54.4 & 87.2  \\
\midrule
\multicolumn{7}{c}{\textit{Open Vision-Language Models - Large}} \\
\midrule 

Qwen2.5-VL-72B & 74.2 & 68.2 & 46.2 & 70.8 & 55.7 & - \\
InternVL2.5-78B &72.3& 70.0 & 48.6 & 69.5 & 50.8 & 88.3\\
InternVL3-78B &79.6& 72.2 & - & 72.5 & 55.9 &89.7\\
VL-Rethinker-32B & 78.8 &65.6  &50.6 & - & -& - \\
VL-Rethinker-72B& 80.4 &68.8  &55.9 & - & -& - \\
\midrule
\multicolumn{7}{c}{\textit{Open Vision-Language Models - Small}} \\
\midrule
Qwen2-VL-7B&58.2 & 54.1 & 30.5 & 60.7 & 33.3& 83.0 \\
Qwen2.5-VL-7B&67.4  &\textbf{58.6}  & 38.3 &62.8  & 42.6&88.3 \\
InternVL2-8B&58.3 & 51.2&29.0 & 62.0 & 33.6& 83.3 \\
InternVL2.5-8B&64.4 &56.0&34.3 & 63.2 & 36.4 & 84.8 \\
Llava-OV-7B&63.2 &48.8&24.1& 61.7 & 33.3& 80.0 \\
\midrule
\multicolumn{7}{c}{\textit{Vision-Language Reasoning Models}} \\
\midrule
Ovis2-8B & 71.8& 57.4 & - & 64.6 & 39.4 & -\\
MiniCPM-V2.6 & 60.6 &49.8 & 27.2 & 60.4& 27.5 & 82.4\\
LLaVA-Next-34B & 46.5 & 51.1& 23.8 & 52.1 & 33.7 & 67.6\\
MM-Eureka-7B & 73.0 & 52.7 &36.3  &62.9  &46.2 &89.0 \\
Vision-R1&73.5  & 49.4 &35.2  &56.0  &44.0&89.9 \\
ThinkLite-VL&\underline{75.1}  & 53.6 &40.1  &\underline{63.0}  &\underline{48.0}&\underline{90.5} \\
VL-Rethinker-7B& 74.9 &\underline{56.7}  &\textbf{41.7}  & 61.9 & 46.6&\underline{90.5} \\

\midrule
\midrule
\textbf{\model (ours)}& \textbf{76.1} & 53.4 & \underline{41.2}  & \textbf{66.0} & \textbf{50.2} & \textbf{90.8}   \\
$\Delta_{\text{Qwen2.5-VL-7B}}$ & {\scriptsize $(+8.7)$} 
 & {\scriptsize $(-5.2)$} 
 & {\scriptsize $(+2.9)$} 
 & {\scriptsize $(+3.2)$} 
 & {\scriptsize $(+7.6)$} 
 & {\scriptsize $(+2.5)$} \\

\bottomrule
\label{tab:comprehensive}
\end{tabular}
}
\label{tab:general}
\end{table}

\paragraph{Difficulty-based Data Filtering.}
The extremely easy or extremely hard data instances result in zero advantage in Eq. \ref{eq:grpo}, thus making no contribution to training \citep{yu2025dapo}. In our initial experiments, we also found that even for data instances that yield non-zero advantage, their contributions are either trivial (for easy instances) or harmful (for difficult instances). Specifically, we observe that for difficult instances, the model’s “correct” rollout is still incorrect. Since the rule-based reward model only verifies the final result and cannot evaluate intermediate steps, which is usually wrong from our observation.
Therefore, we filter data instances based on their difficulty relative to the training model. Concretely, for each instance, we use the checkpoint from the previous training round to perform $k$ rollouts and compute the average accuracy. We retain only those instances with an average accuracy between 0.2 and 0.8 (inclusive) for RL training.

\paragraph{Multi-round RL Training.}
In the multi-round training process, we iterate the above difficulty-based data filtering and model training until convergence.
In each round, we utilize our checkpoint in the last round to estimate the difficulty of the untrained data, and then select moderate difficult samples for training.
For RL training, we adopt the Group Relative Policy Optimization~(GRPO) algorithm~\citep{shao2024deepseekmath}, and stop training once the reward score and validation set results converge. More details of the RL training are shown in Appendix \ref{app-sec:rlvl}.

\begin{table}[t]
\centering
\small
\caption{Benchmarking results for mathematical reasoning tasks. The best and second-best scores for 7B models on each benchmark are shown in \textbf{bold} and \underline{underline}, respectively. \textbf{Avg.} denotes the average performance over the five mathematical benchmarks. For MathVision and MathVerse, we report the numbers on the mini testset. For WeMath, the results presented are from the \texttt{strict} criteria.}

\resizebox{\textwidth}{!}{
\begin{tabular}{lccccccc}
\toprule
\textbf{Models} & \textbf{MathVision} & \textbf{MathVerse} & \textbf{OlympiadBench} & \textbf{WeMath} & \textbf{DynaMath} & \textbf{Avg.}\\ 
\midrule
\multicolumn{7}{c}{\textit{Proprietary Vision-Language Models}} \\
\midrule
GPT-4o & 30.6  &47.8  &25.9  &  50.6 & 63.7 & 45.7 \\
Claude-3.5 & 33.5 &41.2 &- &- & 64.8& -\\
Gemini-1.5 Pro & 19.2 & 54.8 & -& 26.4 & 60.5 & -\\
\midrule
\multicolumn{7}{c}{\textit{Open Vision-Language Models - Large}} \\
\midrule
Qwen2.5-VL-72B & 38.1 & 57.6 & 30.2 & 49.1 & 67.1 & 48.4 \\
InternVL2.5-78B & 32.2 & 51.7 & 11.6 & 39.8 & 19.2 & 28.4 \\
InternVL3-78B & 43.1 & 51.0 & 44.6 & 46.1 & 35.1 & 44.0 \\
\midrule
\multicolumn{7}{c}{\textit{Open Vision-Language Models - Small}} \\
\midrule
Qwen2.5-VL-7B&25.1   &46.3  & 20.2 & 36.2 &55.6 & 36.7  \\
InternVL2.5-8B&19.7 & 39.5 & 12.9 & 23.5 & 39.1& 26.9\\
\midrule
\multicolumn{7}{c}{\textit{Vision-Language Reasoning Models}} \\
\midrule
MM-Eureka-7B & 26.9 &50.3  &20.1  &34.9&56.3& 37.7  \\
Vision-R1-7B& \textbf{32.3}  &\underline{52.4}  &21.1  & \textbf{50.5}&56.0 & \textbf{42.5}  \\
R1-VL-7B&24.7 & 40.0&12.1  &-  &45.8&-\\
R1-Onevision-7B& 29.9  & 46.4 &16.9  &30.0  & 53.1 &35.3\\
OpenVLThinker-7B& 25.3  & 47.9 &19.5  &36.9  &55.0 & 36.9\\
ThinkLite-VL-7B &28.1  &50.7  &22.3  & 41.6&55.9 & 39.7 \\
VL-Rethinker-7B&\textbf{32.3}  & \textbf{54.2} & \textbf{24.0} &  41.7&\underline{57.1} & 41.9\\

\midrule
\midrule
\textbf{Vision-G1 (ours)}& \underline{31.3}  & 51.9 & \underline{23.7} & \underline{45.1}  & \textbf{58.5} & \underline{42.1}  \\
$\Delta_{\text{Qwen2.5-VL-7B}}$  & {\scriptsize $(+6.2)$} 
 & {\scriptsize $(+5.6)$} 
 & {\scriptsize $(+3.5)$} 
 & {\scriptsize $(+8.9)$} 
 & {\scriptsize $(+2.9)$} 
 & {\scriptsize $(+5.4)$} \\
\bottomrule
\end{tabular}
}

\label{tab:math-related}
\end{table}

\section{Experiments}
In this section, we describe the experimental setup and present the results for \model. We first outline the implementation details, the evaluation benchmarks, and the baseline methods (Section \ref{sec:exp_setup}). The primary results comparing \model with the baselines are reported in Section \ref{sec:main_results}, demonstrating the state-of-the-art performance achieved by our model. Furthermore, we provide additional analyses in Section \ref{sec:further_analysis} to examine the effectiveness of our proposed approach.

\subsection{Experimental Setup} \label{sec:exp_setup}

\paragraph{Implementation Details.}
Following the dataset construction pipeline in Section~\ref{sec:data_build}, we create a comprehensive and high-quality RL-ready training dataset with verifiable reward to train our \model. 
Math-related problems constitute half of the training dataset; the remaining domains (\eg chart and medical problems) are uniformly represented. After applying the influence-function-based filtering to remove low-quality data items, the final training set contains 40k questions. When estimating the instance difficulty, we set $k=16$ to balance efficiency and performance.
For RL training, we use an efficient framework \texttt{verl}\footnote{\url{https://github.com/volcengine/verl}} to implement the GRPO algorithm.
We initialize the model weights with Qwen2.5-VL-7B-Instruct~\citep{bai2025qwen25vl} and train it for two rounds. The batch size is set to 128. For each question in a batch, we randomly sample 32 responses from the model as the rollout results and use the answer accuracy as the reward for each response. The model is trained with $8\times$ NVIDIA H200 GPUs for around 18 hours.
For answer accuracy computation, we use the open source tool \texttt{math-verify}\footnote{\url{https://github.com/huggingface/Math-Verify}} in conjunction with normalized exact string matching to compare the ground truth with the model-predicted answer. The reward score range is $[0.0, 1.0]$. 
During evaluation, we use greedy decoding to generate a single response for each question in the benchmark. Accuracy is computed with the same answer-matching protocol as in training, and the model performance is reported as \textbf{Pass@1} unless otherwise specified.

\begin{table}[tbp]
\centering
\small
\caption{Benchmarking results for domain-specific reasoning tasks, including infographics (charts), medical, and cross-image reasoning. The best and second-best scores for each benchmark are shown in \textbf{bold} and \underline{underline}, respectively.}
\begin{tabular}{lccccccccc}
\toprule
\multirow{2}{*}{\textbf{Models}}  & \textbf{Charxiv}& \textbf{ChartQA} & \textbf{VQA} & \textbf{Path} & \textbf{SLAKE} & \textbf{Muir} \\ 
&\textbf{(R/D)}&\textbf{-Pro}&\textbf{-RAD}&\textbf{-VQA}&&\textbf{-Bench}&\\
\midrule
GPT-4o &47.1/84.4  & 41.7 & - & - &- & 68.0 \\
Claude-3.5 &60.2/84.3 & 53.7 &-  & - & - & -\\
Gemini-1.5 Flash &33.9/- & 46.0 &-  & - & - & -\\
Gemini-1.5 Pro &43.3 / 72.0  & - &-  & - & - & -\\
\midrule
Qwen2.5-VL-7B&42.7/\textbf{73.5}  &\underline{46.7}  & \textbf{74.5} & 65.2 & 76.3 &39.8 \\
MM-Eureka-7B&41.3/67.8  &39.9  & 40.2  & 46.1  & 57.2 &23.9 \\
Vision-R1&38.9/57.6  &39.7  & 64.9 & 48.5 & 65.1 &41.3 \\
ThinkLite-VL& \underline{43.8}/65.0 &46.2  &70.5  & \textbf{68.1} & \textbf{78.6} &59.0 \\
VL-Rethinker&42.8/\underline{69.3}  &41.5  & 56.2 & 66.1 & 59.7 &58.3 \\
\midrule
\textbf{\model (ours)}& \textbf{44.0}/65.5 & \textbf{47.7} & \underline{72.1} & \underline{66.7} & \underline{78.3} & \textbf{61.5} & \\
$\Delta_{\text{Qwen2.5-VL-7B}}$   & {\scriptsize $(+1.3)/(-8.0)$}
 & {\scriptsize $(+1.0)$}
 & {\scriptsize $(-2.4)$}
 & {\scriptsize $(+1.5)$}
 & {\scriptsize $(+2.0)$}
 & {\scriptsize $(+21.7)$} \\

\bottomrule
\label{tab:domain-specific}
\end{tabular}

\label{tab:main}
\end{table}

\paragraph{Evaluation Benchmarks.}
We evaluate our model on a set of comprehensive visual reasoning benchmarks, including MathVista~\citep{lu2024mathvista}, MMMU-Val~\citep{yue2024mmmu}, MMMU-Pro\citep{yue2024mmmupro}, and MMStar \citep{chen2024wemmstar}.
The four benchmarks comprehensively evaluate the visual reasoning abilities of VLMs from multiple dimensions, covering visual puzzles, college-level problems, and science questions.
Additionally, we evaluate on LogicVista \citep{xiao2024logicvista} and ChartQA \citep{masry2022chartqa}; while they emphasize broad logical reasoning and chart/plot understanding, respectively, their scope is sufficiently comprehensive that we also classify them as general visual reasoning benchmarks.
For mathematical visual reasoning, we include 5 widely-used benchmarks: MathVision \citep{wang2024mathvision}, MathVerse \citep{zhang2024mathverse}, OlympiadBench \citep{he2024olympiadbench}, WeMath \citep{qiao2024wemath}, and DynaMath \citep{zou2024dynamath}. 
These benchmarks contain math problems that require image understanding to solve, covering skills from simple counting and perceptual reasoning to complex geometry and combinatorial reasoning.
To target specific domains, we assess chart and plot reasoning with ChartXiv \citep{wang2024charxiv} and ChartQAPro \citep{masry2025chartqapro}, and medical visual reasoning with VQA-RAD \citep{vqarad}, PathVQA \citep{he2020pathvqa}, and SLAKE \citep{liu2021slake}.
For multi-image reasoning, we choose MuirBench \citep{wang2024muirbench}, which contains 12 multi-image understanding tasks.

\paragraph{Baseline Methods.}
To comprehensively verify the effectiveness of our method, we mainly compare it against VLMs with a similar parameter scale.
Specifically, we first select five VLMs with around 7B size, including Qwen2.5-VL-7B \citep{bai2025qwen25vl}, Ovis-8B~\citep{lu2024ovis}, MiniCPM-V2.6 \citep{yao2024minicpm}, Llava-OV \citep{li2024llava-ov}, and LLaVA-Next~\citep{llava_next}.
Among all the above models, Qwen2.5-VL-7B generally performs the best and has been widely used in existing reasoning VLM work as the backbone.
In addition, we also considered the following set of recently proposed reasoning-oriented VLMs that have incorporated Reinforcement Learning (RL) during training, \ie MM-Eureka-7B \citep{meng2025mm}, Vision-R1-7B \citep{huang2025visionr1}, ThinkLite-VL-7B \citep{wang2025thinklite}, and VL-Rethinker-7B \citep{wang2025vlrethinker}. 
All the four reasoning-focused models adopt RL as a core component to improve multimodal reasoning capabilities. Concretely, MM-Eureka-7B follows a hybrid paradigm combining supervised fine-tuning (SFT) with subsequent RL training to refine reasoning behaviors. %
OpenVLThinker-7B and ThinkLite-VL-7B employ iterative self-improvement pipelines, leveraging reasoning traces from earlier model outputs. 
Finally, MM-Eureka-7B and Vision-R1-7B further integrate custom reward mechanisms or rule-based guidance, while VL-Rethinker-7B and ThinkLite-VL-7B introduce strategies such as forced rethinking and MCTS-guided selection to promote deeper, data-efficient reasoning.
Note that all the above baselines utilize QWen2.5-VL-7B as the backbone to perform RL training, including our method \model.
To contextualize the performance of our method, we also report the results from several state-of-the-art large VLMs and closed-source products as a reference, \ie GPT-4o\footnote{\url{https://chatgpt.com/}}, Claude-3.5-Sonnet\footnote{\url{https://claude.ai/}}, Gemini-1.5-Flash\footnote{\url{https://deepmind.google/technologies/gemini}}, Gemini-1.5-Pro, Qwen2.5-VL-72B, InternVL2.5-78B~\citep{chen2024expanding}, and InternVL3-78B~\citep{zhu2025internvl3}.

\subsection{Main Results} \label{sec:main_results}

We conduct extensive experiments on 18 benchmarks and discuss the model's performance on the following three types of visual reasoning tasks.

\paragraph{Evaluation on Comprehensive Visual Reasoning Tasks.}
As shown in Table~\ref{tab:comprehensive}, 7B-scale VLMs trained with RL substantially outperform the base model, \ie Qwen2.5-VL-7B-Instruct, highlighting the effectiveness of RL in eliciting the visual reasoning ability of VLMs.
Among all the RL-trained methods, ThinkLite-VL performs well in four benchmarks (\ie MathVista, MMStar, LogicVista and ChartQA).
ThinkLite-VL adopts a MCTS-guided sample-selection method that estimates instance difficulty over multiple iterations, suggesting that appropriate data filtering strategy can significantly benefit RL training for VLMs.
Benefiting from both RL training and the carefully designed data arrangement methods, our method achieves the best performance on most benchmarks, achieving 1.6\% absolute improvement on average.
Moreover, despite not explicitly training on logical-reasoning datasets, our model performs competitively on LogicVista, indicating that it has learned transferable and generalizable reasoning skills from other data sources.
Finally, our approach can achieve comparable or even better performance than larger VLMs and proprietary models, \eg InvernVL2.5-78B and Gemini-1.5 Flash, highlighting the efficacy of reinforcement learning when coupled with well-curated data.

\begin{table}[t]
\centering
\small
\caption{Ablation study results for comprehensive visual reasoning tasks. The best and second-best ones are marked in \textbf{bold} and \underline{underlined}, respectively.}
\label{tab:ablation}
\begin{tabular}{lcccccccc}
\toprule
\textbf{Models}
  & \textbf{\makecell{Math\\Vista}}
  & \textbf{\makecell{Math\\Vision}}
  & \textbf{\makecell{MMStar}}
  & \textbf{\makecell{Logic\\Vista}}
  & \textbf{\makecell{ChartQA\\Pro}}
  & \textbf{\makecell{VQA\\-RAD}}
  & \textbf{\makecell{Muir\\Bench}} \\
\midrule
Qwen2.5-VL-7B & 67.4 & 25.1 & 62.8 & 42.6 & 46.7 & \textbf{74.5} & 39.8 \\
\midrule
\textbf{\model (ours)}  &\underline{76.1} & \textbf{31.3} & \textbf{66.0} &\textbf{50.2} & \underline{47.7}& \underline{72.1} & \textbf{61.5} \\
\quad w/o Multi-round  &74.9 &29.6 & 64.8 & 46.0&\textbf{48.4}&\underline{72.1}& 57.7& \\
\quad w/o Data Selection & 71.5 &  25.3 & 64.2 & 44.0& 42.9 &69.7& \underline{59.4} \\
\quad    w/o Domain-specific Datasets &\textbf{76.3} &\underline{30.3} &\underline{65.3} &\underline{48.0} &44.5 & 68.1 & 58.5 \\
\bottomrule
\end{tabular}
\end{table}

\paragraph{Evaluation on Math-Related Visual Reasoning Tasks.}
Table \ref{tab:math-related} reports results of math-related visual reasoning tasks. Our proposed \model, along with all RL-trained baselines, substantially improves over the base model.
On average, our \model achieves the best performance, underscoring the effectiveness of our curated math datasets and training recipe. We also observe that Vision-R1 performs strongly on MathVision, MathVerse, and WeMath benchmarks; we attribute this to its focused mathematical training datasets, which emphasizes math-specific objectives. 
However, this specialization transfers less effectively to datasets in other domains, such as MMMU and MMStar, where its performance lags. In contrast, our \model achieves higher average accuracy than Vision-R1 and generalizes more reliably across domains, suggesting a better balance of visual reasoning skills and making it a better general reasoning VLM.
In addition, it is worth noting that our model's performance on MathVista \citep{lu2024mathvista} is even better than OpenAI's o1 \citep{OpenAIo1}.

\paragraph{Evaluation on Domain-specific Reasoning Tasks.}
As shown in Table~\ref{tab:domain-specific}, several RL-trained VLMs exhibit performance degradation relative to the base model on diverse domain-specific benchmarks, especially in domains underrepresented in their RL data.
In contrast, our method demonstrates stronger robustness and generally better results in these benchmarks.
We can attribute this to three design choices: (i) influence-function–based filtering that removes instances estimated to be harmful to other training datasets or downstream tasks; (ii) balanced sampling across tasks and domains to avoid overfitting to any single domain; and (iii) a multi-round RL schedule with difficulty-based data selection, which stabilizes training and promotes progressive learning.

\begin{figure}[t]
  \centering
  \includegraphics[width=1.0\textwidth]{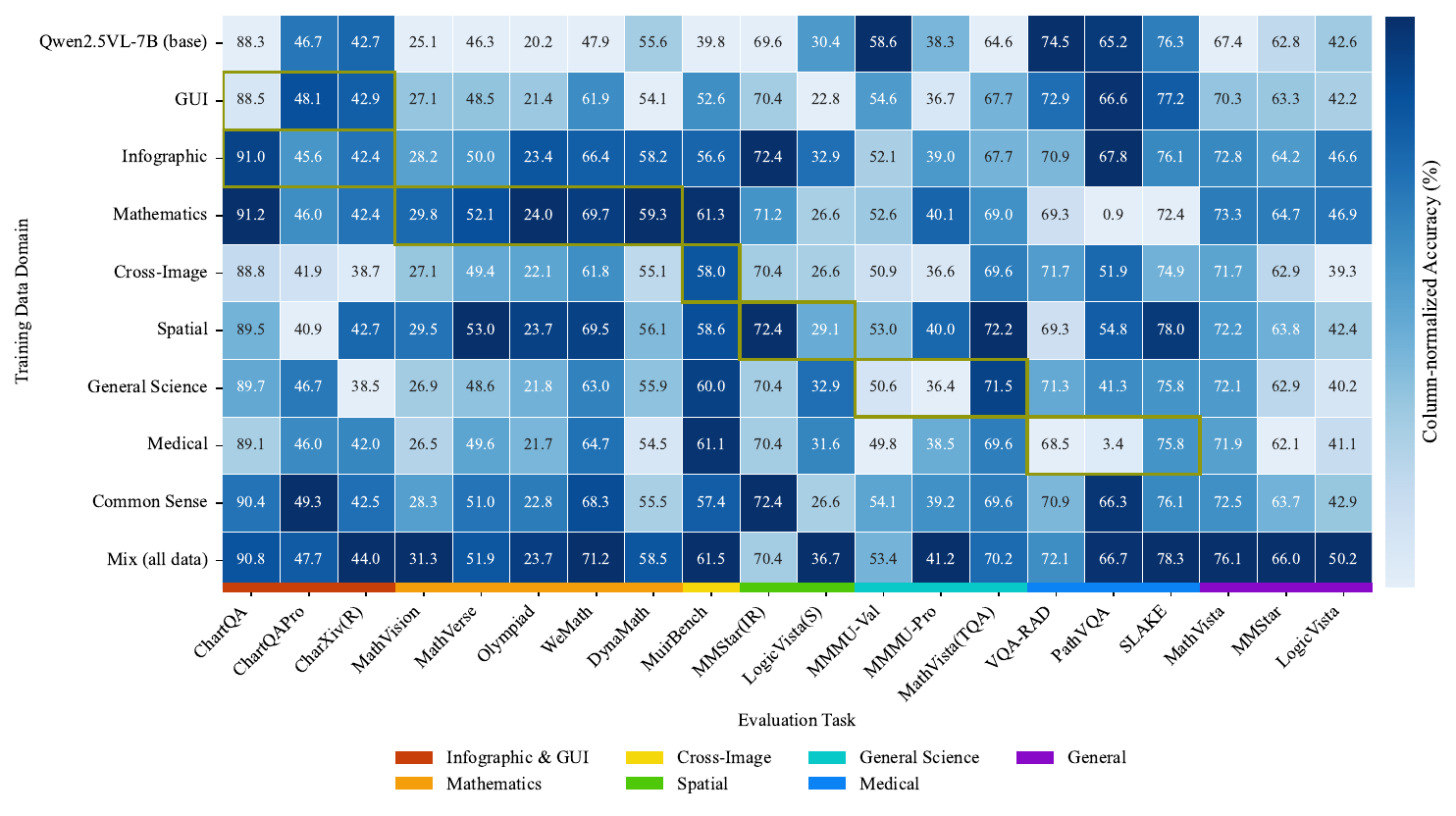}
  \caption{Heatmap illustrating the contribution of each data domain. Darker \textcolor[HTML]{00008B}{\textbf{blue}} represent higher performance, while lighter \textcolor[HTML]{ADD8E6}{\textbf{blue}} indicate lower performance. Non-blue color legends denote the domains of the evaluation tasks (benchmarks). The blocks bounded by \textcolor[HTML]{868D27}{\textbf{olive drab}} boxes represent in-domain benchmarking results. We extract and evaluate the \emph{Instance Reasoning} subset of MMStar (abbrev.\ MMStar(IR)) and the \emph{Spatial} subset of LogicVista (abbrev.\ LogicVista(S))—while excluding other categories. Mixing all domain data with our data selection strategy yields the overall best performance, as shown in the last row.}
  \label{fig:domain_heat_map}
\end{figure}

\subsection{Further Analysis} \label{sec:further_analysis}
We further analyze the model training design, examining the contribution of each data domain to overall performance. The ablation studies and training process analysis show the effectiveness of both the data selection strategy and the multi-round RL training.

\paragraph{Domain Contribution.} We study the contribution from each data domain. Following Section \ref{sec:data_build}, we first categorize the raw training questions into eight reasoning types: graphic user interface (GUI), infographic, mathematics, cross-image, spatial, general science, medical, and common sense. We then randomly sample 10k items from each domain to train the base model. For efficiency, we choose Qwen2.5VL-7B-Instruct as the base model. We conduct just one round training without difficulty filtering. The result in Figure \ref{fig:domain_heat_map} shows that the mixed-domain training with influence-function–based selection plus difficulty-based filtering yields the strongest overall performance, outperforming any single-domain model. Besides, capabilities acquired from mathematical data generalize to other domains—most notably infographics and cross-image reasoning—though transfer is not universal across all targets; conversely, training on infographic data exhibits positive transfer to mathematical benchmarks. We also notice that models trained solely on medical data underperform even on medical benchmarks, highlighting the necessity of mixing domains to facilitate learning.

\paragraph{Ablation Study.}
In this part, we conduct ablation studies to assess the contributions of our multi-round RL training and data selection design.
Concretely, we test the following variations of our method: (i) \emph{w/o Multi-round} performs one-round RL training until convergence; (ii) \emph{w/o Data Selection} randomly selects 40k (matching our main training size) instances from the raw dataset without selection and filtering; (iii) \emph{w/o Domain-specific Datasets} trains only on two high-quality datasets (\ie ThinkLite and ViRL39k).
As shown in Table~\ref{tab:ablation}, each ablation underperforms our method, indicating that all our design components are necessary for the observed gains.
Besides, the variation w/o Domain-specific Datasets can achieve better performance on MathVista, but degrades substantially on most domain-specific datasets.
It further proves that adding a variety of domain-specific training datasets is important for VLMs to develop general reasoning capability.

\begin{figure}[ht]
  \centering
  \includegraphics[width=0.90\linewidth]{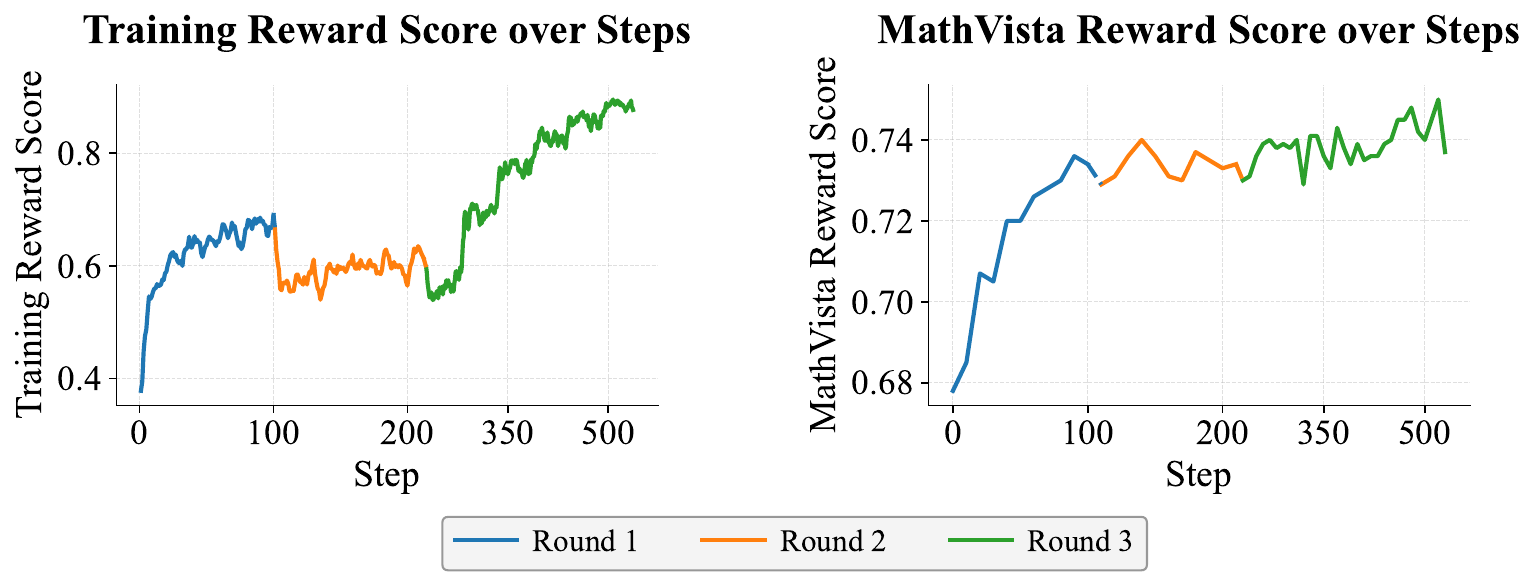}
  \caption{The progress in the multi-round RL training. \textbf{Left:} The mean reward score over steps. \textbf{Right:} The accuracy score on MathVista-mini over steps. Due to resource limitation, we only compute the accuracy with the rule-based reward function on MathVista-mini. Therefore, the scores (curve) are lower than the final evaluation, which uses GPT-4o-mini as an additional judge.}
  \label{fig:retrieval_evaluation}
\end{figure}

\paragraph{Training Process Analysis.}
Our multi-round RL training procedure, combined with influence-aware filtering and difficulty-based selection, yields a stable training trajectory with steadily improving capability. 
To verify the training stability and effectiveness, we log the mean reward scores and the MathVista-Mini test scores at 10-step intervals throughout all rounds' training process.
As shown in Fig.~\ref{fig:retrieval_evaluation}, the first-round training exhibits rapid reward growth, and the performance on MathVista also starts to converge.
Before the second round, we reconstruct the training set using the difficulty-based filter to emphasize moderately challenging instances, which induces a brief reward drop around step 100, followed by consistent gains in both reward and held-out accuracy. We repeat the steps to get a filtered dataset for round three, which subsequently shows a similar trend over the training reward and MathVista performance.

\section{Conclusion}
In this paper, we present a general reasoning VLM, \model, trained via reinforcement learning.
To achieve this, we construct a large RL-ready dataset by assembling 46 tasks across 13 dimensions within 8 domains and unifying the data format.
We then devise an influence function-based filtering strategy to remove low-quality instances.
After filtering, we perform multi-round RL training using GRPO, alternating difficulty-based data selection with training to gradually enhance general reasoning ability.
In each round, we ensure the intermediate dataset contains moderately difficult instances with a balanced category distribution.
Experiments on 17 benchmarks demonstrate the effectiveness of our method, surpassing state-of-the-art models of similar scale and even outperforming GPT-4o and Gemini-1.5.
In the future, we will extend our method to more real-world tasks and scenarios, such as video understanding and 3D perception, and explore on-policy data synthesis methods for automatically generating new high-value instances.
We expect our corpus, data preparation framework, and open-source checkpoints to serve as a foundation for the next generation of reasoning-centric VLM research.

\bibliographystyle{unsrt}
\bibliography{ref.bib}

\appendix

\newpage

\clearpage
\appendix
\section{Implementation Details of \model} 
\subsection{Reinforcement Learning for Vision Language Model} \label{app-sec:rlvl}
To train our reasoning vision language model, \model, we use the Group Relative Policy Optimization (GRPO) \citep{shao2024deepseekmath} algorithm, which is a variant reinforcement learning algorithm of Proximal Policy Optimization (PPO) \citep{schulman2017proximal}. GRPO computes a group-relative advantage from multiple samples of the same prompt instead of learning a seperate value function. It is effective for training problems whose correctness can be automatically checked. During the training, we optimize the VLM with thel GRPO loss $\mathcal{J}_{\mathrm{GRPO}}(\theta)$:
\begin{equation}
\label{eq:grpo}
\begin{aligned}
&\mathcal{J}_{\mathrm{GRPO}}(\theta)
= \mathbb{E}_{\left[q\sim P(Q),\,\{o_i\}_{i=1}^G\sim\pi_{\theta_{\mathrm{old}}}(O\mid q)\right]} \\
&\Biggl[ 
\frac{1}{G}\sum_{i=1}^G\frac{1}{\lvert o_i\rvert}
\sum_{t=1}^{\lvert o_i\rvert}
\Bigl\{
\min\Bigl[
\frac{\pi_{\theta}(o_{i,t}\mid q,o_{i,<t})}
     {\pi_{\theta_{\mathrm{old}}}(o_{i,t}\mid q,o_{i,<t})}
\hat A_{i,t},
\mathrm{clip}\Bigl(
\frac{\pi_{\theta}(o_{i,t}\mid q,o_{i,<t})}
     {\pi_{\theta_{\mathrm{old}}}(o_{i,t}\mid q,o_{i,<t})},
1-\epsilon,1+\epsilon
\Bigr)\hat A_{i,t}
\Bigr] \\
&-
\beta\,\mathbb{D}_{\mathrm{KL}}\!\bigl[\pi_{\theta}\,\Vert\,\pi_{\mathrm{ref}}\bigr]
\Bigr\}
\Biggr],
\end{aligned}
\end{equation}
where $\pi_\theta$ and $\pi_{\theta_{old}}$ are the current and old policy. $\pi_{ref}$ is the reference model, which, in this case, is the Qwen2.5-VL-7B-Instruct model. $q$ and $o$ are the questions and outputs sampled from our dataset and the old policy $\pi_{\theta_{old}}$, respectively. $\epsilon$ and $\beta$ are hyper-parameters for stabilizing training. $\hat{A}_{i,t}$ is the advantage of the relative rewards of the outputs in each group. For each response, we use rule-based method to evaluate whether it is correct or not. The correct response will get a positive reward ($r_i=1$), while the incorrect response will get a zero reward ($r_i=0$). We use the normalized reward as the advantage: $\hat A_{i,t}=\frac{r_i-\text{mean}(\mathbf{r})}{\text{std}(\mathbf{r})}$. We use an unbiased estimator to estimate the KL divergence $\mathbb{D}_{KL}$:
\begin{equation}
    \mathbb{D}_{KL}\bigl[\pi_\theta \,\big\|\, \pi_{ref}\bigr]
= \frac{\pi_{ref}\bigl(o_{i,t}\mid q, o_{i,<t}\bigr)}
       {\pi_\theta\bigl(o_{i,t}\mid q, o_{i,<t}\bigr)}
  - \log\frac{\pi_{ref}\bigl(o_{i,t}\mid q, o_{i,<t}\bigr)}
             {\pi_\theta\bigl(o_{i,t}\mid q, o_{i,<t}\bigr)}
  - 1,
\end{equation}

Following previous work on VLM post-training \citep{bai2025qwen25vl}, We freeze the vision encoder and only tune the language model part of Qwen2.5-VL.

\subsection{Data Selection from Source Datasets}
Training the reasoning VLM using reinforcement learning requires reward for each individual question. Therefore, we use datasets with questions that have verifiable ground truth answers so that a rule-based reward model can be used to estimate their rewards. To build such a training dataset, we collect raw datasets from many domains, including infographic reasoning, mathematical reasoning, cross-image reasoning, spatial reasoning, and knowledge-specific reasoning. The full list of the raw datasets is shown in Table \ref{tab:data_sources}. We also collect four comprehensive visual reasoning datasets: MM‑R1~\citep{peng2025lmmr1}, VerMulti~\citep{peng2025lmmr1}, ThinkLite~\citep{wang2025thinklite}, ViRL39K~\citep{wang2025vlrethinker}, and MMK12~\citep{meng2025mmeureka}.  
ViRL39K and MM-R1 cover a broad range of STEM problems, VerMulti focuses on geometry and diagram-based tasks, and MMK12 comprises K-12–level problems in mathematics, physics, chemistry, and biology.
Note that not all raw data have verifiable ground truth, we use rules to filter out those that do not. Specifically, we use regular expression to find ground truths with numeric values, multiple-choice options, yes/no answers, or other single-word ground truths.

To get the category label for each question in our collected datasets, we use an LLM (Qwen2.5-32B-Instruct) as the judge. To further facilitate data categorization, we split each domain into subcategories, and then prompt the LLM to determine the subcategory of a question. Specifically, we use the prompt in Table \ref{tab:classifier_prompt}.

\begin{table}[ht]
  \centering
  \caption{Training data sources and their corresponding categories. The categories shown here are approximate and serve as a rough classification for each dataset. In our experiments, we perform instance-level classification within these datasets.}
  \begin{tabular}{@{} ccc @{}}
    \toprule
    \multicolumn{3}{c}{\textbf{Training Data Category}} \\
    \cmidrule(lr){1-3}
    \textbf{General Category} & \textbf{Sub-Category} & \textbf{Datasets} \\
    \midrule
    \makecell[c]{Infographic\\Reasoning} & \makecell[c]{Chart/Plot, Table, Map, \\Document, Web} & 
    \makecell[c]{FigureQA\citep{kahou2017figureqa},
    DVQA\citep{kafle2018dvqa},
    PlotQA\citep{methani2020plotqa}, \\
    ChartQA\citep{masry2022chartqa},
    TabMWP\citep{lu2022tabmwp},
    MapQA\citep{chang2022mapqa}, \\
    ChartBench\citep{xu2023chartbench},
    UniChart\citep{masry2023unichart}, \\
    DocVQA\citep{mathew2021docvqa},
    MultiUI\citep{liu2024multiui}} \\
    \midrule
    \makecell[c]{Mathematical\\Reasoning} & \makecell[c]{Geometry, Arithmetic, \\Broader Mathematics} & 
    \makecell[c]{Geometry3K\citep{lu2021geo3k},
    GeoQA+\citep{cao2022geoqa_plus}, \\
    UniGeo\citep{chen2022unigeo}
    GeoQA\citep{chen2021geoqa},
    MMR1\citep{peng2025lmmr1},\\
    GEOS\citep{seo2015geos},
    CLEVR-Math\citep{lindstrom2022clevr_math}} \\
    \midrule
    \makecell[c]{Cross-image\\Reasoning} &  \makecell[c]{Cross-image Reasoning} & 
    \makecell[c]{IconQA\citep{lu2021iconqa},
    NLVR2\citep{suhr2018nlvr2},
    ImageCode\citep{krojer2022imagecode}} \\
    \midrule
    \makecell[c]{Spatial\\Reasoning} & \makecell[c]{Spatial Reasoning} & 
    \makecell[c]{VQA-AS\citep{antol2015vqa},
    Super-CLEVR\citep{li2023super-clevr}} \\
    \midrule
    \makecell[c]{Knowledge-specific\\Reasoning} & \makecell[c]{Science, Medical, \\Common Sense} & 
    \makecell[c]{AI2D\citep{kembhavi2016ai2d},
    TQA\citep{kembhavi2017tqa},
    ScienceQA\citep{lu2022scienceqa},\\
    MMK12\citep{meng2025mmeureka},
    VQA2.0\citep{antol2015vqa}
    VizWiz\citep{gurari2018vizwiz},\\
    TextVQA\citep{singh2019textvqa},
    A-OKVQA\citep{schwenk2022a-okvqa},
    OK-VQA\citep{marino2019ok-vqa}\\
    PMC-VQA\citep{zhang2023pmc-vqa},
    VQA-RAD\citep{vqarad},\\
    SLAKE\citep{liu2021slake},
    Path-VQA\citep{he2020pathvqa}} \\
    \bottomrule
  \label{tab:data_sources}
  \end{tabular}
\end{table}

\ignore{
\begin{table}[ht]
  \centering
  \caption{Training Data Category. We list the correspondence between the general categories and the sub-categories.}
  \begin{tabular}{@{} c @{\hspace{50pt}} cc @{}}
    \toprule
    \multicolumn{3}{c}{\textbf{Training Data Category}} \\
    \cmidrule(lr){1-3}
    \textbf{General Category} & \textbf{Sub-Category} & \textbf{Sub-Category} \\
    \midrule
    \multirow{5}{*}{\makecell[l]{Infographic\\Reasoning}}
      & Chart/Plot \\
      & Table \\
      & Map \\
      & Document \\
      & Web \\
    \midrule
    \multirow{3}{*}{\makecell[l]{Mathematical\\Reasoning}}
      & Geometry \\
      & Arithmetic \\
      & Broader Mathematics \\
    \midrule
    \makecell[l]{Cross-image\\Reasoning}
      & Cross-image Reasoning \\
    \midrule
    \makecell[l]{Spatial\\Reasoning}
      & Spatial Reasoning \\
    \midrule
    \multirow{3}{*}{\makecell[l]{Knowledge\\-specific\\Reasoning}}
      & General Science \\
      & Common Sense \\
      & Medical Specific \\
    \bottomrule
  \label{tab:training_data_category}
  \end{tabular}
\end{table}}

\begin{table}[htbp]
  \centering
  \caption{Prompt used for classifying subcategory of the questions in the raw datasets}
  \begin{tcolorbox}[width=.9\textwidth]
            "You are a QUESTION-TYPE classifier (do **NOT** answer the question itself)."\\\\
        "INSTRUCTIONS"\\
        "Read the question and output ONLY the MOST RELATED SINGLE category name from the list below that CLEARLY apply."\\
        "Make sure the category name is lowercase."\\\\
        "Do NOT explain, do NOT repeat the prompt."\\
        "CATEGORIES"\\
        "chartplot - Bar chart, line chart, pie chart, scatter plot, etc."\\
        "table - Grid of cells with headers for rows and/or columns."\\
        "map - Contains landmarks, coordinates, compass, scale. Represents physical space or layout."\\
        "document - Visuals that contain structured information presented in formats like forms or organized text blocks. Requires extracting or reasoning over layout and textual structure."\\
        "web - Visuals captured from websites, involving multi-modal web elements like buttons, charts, or embedded tables. Tasks focus on interpreting online content."\\
        "geometry - Mathematical geometry problems involving coordinate systems, distances, and equations of lines/curves. Area, perimeter, or segment length computations of shapes like triangles, rectangles, and circles. Volume and surface area of 3D shapes like cubes, spheres, and cones. Understanding vertices, edges, paths, and connections in graphs."\\
        "arithmetic - Basic operations: counting, addition, subtraction, multiplication, division, order of operations."\\
        "broader mathematics - Questions about statistics, continuity, boundaries, or connectedness. Puzzle solving, such as identifying visual or numerical patterns in shapes or sequences, including transformations, symmetries, and progressions."\\
        "crossimage reasoning - Tasks that require comparing or synthesizing information across multiple images. Can involve visual entailment, spotting differences, or deducing outcomes from a series of different image inputs."\\
        "spatial reasoning - Understanding positions, directions, and relationships between objects in real 3D space."\\
        "general science - Questions based on scientific knowledge from physics, chemistry, biology, and related disciplines. Tasks may include interpreting experiments, applying principles, or identifying cause-effect relationships."\\
        "common sense - Requires intuitive world knowledge or social understanding. Tasks often rely on daily experiences, social norms, or expected behaviors."\\
        "medical specific - Requires applying professional medical knowledge to interpret symptoms, test results, or patient history and make diagnostic or treatment decisions. These tasks involve moderate reasoning and contextual judgment."\\
        "question: \{question\}"\\
        "answer:"
  \label{tab:classifier_prompt}
  \end{tcolorbox}
\end{table}

\subsection{Format Unification} \label{app:format}
The visual reasoning problems collected from various data sources come in different formats. To facilitate RL training, we standardize their formats. Specifically, we retain only the following items.
\begin{itemize}
    \item \textbf{Prompt.} The prompt contains both the original question and the user instruction. We append the user instruction after each original question following Thinklite \citep{wang2025thinklite}. For example, "\texttt{What is the ratio between the two top bars? You FIRST think about the reasoning process as an internal monologue and then provide the final answer. The reasoning process MUST BE enclosed within <think> </think> tags. The final answer MUST BE put in \textbackslash boxed\{\}.}"
    \item \textbf{Images.} A list of image bytes.
    \item \textbf{Ground Truth.} A numeric value or string representing the ground truth answer to the question, such as \texttt{1.06}.
\end{itemize}

\section{Additional Experiment Results}
\subsection{Benchmarking Result Source}
We collect the performance of the baseline models from various sources. Specifically, we first collect the results from the official benchmarks reported in the paper, e.g., MathVista \citep{lu2024mathvista}. Then, We obtain the results from the papers that propose the model, including VL-Rethinker \citep{wang2025vlrethinker}, InternVL2.5 \citep{chen2024expanding}, InternVL3 \citep{zhu2025internvl3}, Qwen2-VL \citep{qwen2vl}, Qwen2.5-VL \citep{bai2025qwen25vl}, Llava-OneVision \citep{li2024llava-ov}, Llava-Next \citep{llava_next}, Kimi-VL \citep{team2025kimi}, Ovis2 \citep{lu2024ovis}, MiniCPM-v2.6 \citep{yao2024minicpm}, MM-Eureka \citep{meng2025mmeureka}, Vision-R1 \citep{huang2025visionr1}, ThinkLite-VL \citep{wang2025thinklite}, VL-Rethinker \citep{wang2025vlrethinker}, Chart-R1 \citep{chen2025chart}, VLAA-Thinking \citep{chen2025vlaathinker}. When official numbers are unavailable or incomplete, we additionally consult the Open LMM Reasoning Leaderboard and the Open VLM Leaderboard provided by VLMEvalKit \citep{duan2024vlmevalkit} to cross-check scoring and fill gaps. Finally, we reproduce the missing results to compare with our model.

\section{Case Study of \model's Output Reasoning Process}
Our \model shows significant improvement on its reasoning capability in various domains (Section \ref{sec:main_results}). It would be interesting to investigate the change in the reasoning patterns that cause the performance gain. We show a few cases in Table \ref{tab:case-1}, \ref{tab:case-2}, and \ref{tab:case-3}. For example, \model will try to decompose the calculating process into more fine-grained styles, making it more robust when solving complex calculation problems (Table \ref{tab:case-1}). Our model also shows better perception capability when counting objects in the image (Table \ref{tab:case-2}). When solving problems like finding maximums, \model will list more possibilities than the base model, and find the optimal solution at the end (Table \ref{tab:case-3}).

\begin{table}[h]
\centering
\caption{Case Study: Question 1. \boxed{} denotes a special symbol for helping extract the answer.}
\label{tab:case-1}
\begin{tabular}{@{}p{2cm} p{12cm} @{}}
\toprule
   & \includegraphics[width=6cm]{case_study/980.png}  \\
Question & 	<image>What is the sum of 2002, 2003 and 2004? You FIRST think about the reasoning process as an internal monologue and then provide the final answer. The reasoning process MUST BE enclosed within <think> </think> tags. The final answer MUST BE put in \boxed{}.\\ \midrule
Ground Truth & 70.4 \\ \midrule
Base Model  & To find the sum of the internet user share percentages for the years 2002, 2003, and 2004, I need to identify the values for each of these years from the graph:

- For 2002, the value is 19.4\%.
- For 2003, the value is 23.2\%.
- For 2004, the value is 27.8\%.

Now, I will add these values together:

19.4\% + 23.2\% + 27.8\% = 60.4\%

The final answer is \boxed{60.4}.  \\ \midrule
\model    & 
Let's calculate the sum step by step:

1. Identify the values for each year:
   - 2002: 19.4\%
   - 2003: 23.2\%
   - 2004: 27.8\%

2. Add these values together:
   \[
   19.4 + 23.2 + 27.8
   \]

3. First, add 19.4 and 23.2:
   \[
   19.4 + 23.2 = 42.6
   \]

4. Next, add the result to 27.8:
   \[
   42.6 + 27.8 = 70.4
   \]

So, the sum of 2002, 2003, and 2004 is \boxed{70.4}.
\\ \bottomrule
\end{tabular}
\end{table}

\begin{table}[h]
\centering
\caption{Case Study: Question 2. \boxed{} denotes a special symbol for helping extract the answer.}
\begin{tabular}{@{}p{2cm} p{12cm} @{}}
\toprule
   & \includegraphics[width=5cm]{case_study/112.png}  \\
Question & 	<image>How many baseballs are there? You FIRST think about the reasoning process as an internal monologue and then provide the final answer. The reasoning process MUST BE enclosed within <think> </think> tags. The final answer MUST BE put in \boxed{}.\\ \midrule
Ground Truth & 20 \\ \midrule
Base Model  & To determine the total number of baseballs, I'll count the number of baseballs in one bucket and then multiply by the number of buckets.

1. Count the baseballs in one bucket:
   - There are 6 baseballs in one bucket.

2. Count the number of buckets:
   - There are 4 buckets in total.

3. Multiply the number of baseballs in one bucket by the number of buckets:
   - 6 baseballs/bucket * 4 buckets = 24 baseballs

Therefore, the total number of baseballs is 24.

\boxed{24} \\ \midrule
\model    & Let's count the baseballs in each bucket:

1. The first bucket has 5 baseballs.
2. The second bucket has 5 baseballs.
3. The third bucket has 5 baseballs.
4. The fourth bucket has 5 baseballs.

Now, let's add them up:

\[ 5 + 5 + 5 + 5 = 20 \]

Let's count:
- Bucket 1: 5
- Bucket 2: 5
- Bucket 3: 5
- Bucket 4: 5

\boxed{20}

\\ \bottomrule
\label{tab:case-2}
\end{tabular}
\end{table}

\begin{table}[h]
\centering
\caption{Case Study: Question 3. \boxed{} denotes a special symbol for helping extract the answer.}
\label{tab:case-3}
\begin{tabular}{@{}p{2cm} p{12cm} @{}}
\toprule
   & \includegraphics[width=5cm]{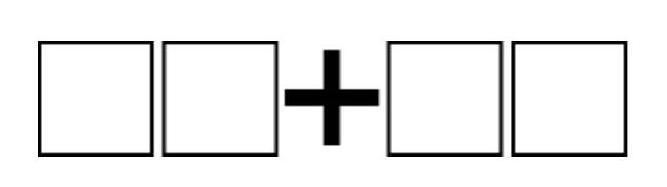}  \\
Question & <image>Each of the digits 2, 3, 4 and 5 will be placed in a square. Then there will be two numbers, which will be added together. What is the biggest number that they could make? <image1> You FIRST think about the reasoning process as an internal monologue and then provide the final answer. The reasoning process MUST BE enclosed within <think> </think> tags. The final answer MUST BE put in \boxed{}. \\ \midrule
Ground Truth & 95\\  \midrule
Base Model  & To solve this problem, we need to maximize the sum of two two-digit numbers formed by the digits 2, 3, 4, and 5. Let's break it down step by step.

1. **Identify the digits and their possible placements:**
   - We have the digits 2, 3, 4, and 5.
   - We need to form two two-digit numbers.
   - The digits must be used exactly once.

2. **Maximize the first number:**
   - To maximize the first number, we should place the largest digits in the tens place.
   - The largest digit is 5, so we place 5 in the tens place of the first number.
   - The next largest digit is 4, so we place 4 in the tens place of the second number.
   - Now we have the digits 2 and 3 left.

3. **Maximize the second number:**
   - To maximize the second number, we should place the next largest digits in the tens place.
   - The next largest digit is 3, so we place 3 in the tens place of the second number.
   - The remaining digit is 2, so we place 2 in the tens place of the first number.

4. **Form the numbers:**
   - The first number is 52.
   - The second number is 34.

5. **Calculate the sum:**
   - Sum = 52 + 34 = 86.

Therefore, the biggest number that they could make is \boxed{86}. \\ \midrule
\model & To determine the largest possible sum, we need to place the digits 2, 3, 4, and 5 in the squares such that the two numbers are maximized. Let's denote the two numbers as \(AB\) and \(CD\), where \(A, B, C,\) and \(D\) are the digits 2, 3, 4, and 5.

We want to maximize the expression \(10A + B + 10C + D\). To do this, we should assign the largest digits to the tens place of each number. Let's try the following arrangement: \(A = 5\), \(B = 4\), \(C = 3\), and \(D = 2\). This gives us the numbers 54 and 32.

Now, let's calculate the sum:
\[54 + 32 = 86\]

To ensure this is the maximum, we should consider other permutations. If we try \(A = 5\), \(B = 3\), \(C = 4\), and \(D = 2\), we get the numbers 53 and 42:
\[53 + 42 = 95\]

If we try \(A = 5\), \(B = 2\), \(C = 4\), and \(D = 3\), we get the numbers 52 and 43:
\[52 + 43 = 95\]

If we try \(A = 4\), \(B = 5\), \(C = 3\), and \(D = 2\), we get the numbers 45 and 32:
\[45 + 32 = 77\]

If we try \(A = 4\), \(B = 3\), \(C = 5\), and \(D = 2\), we get the numbers 43 and 52:
\[43 + 52 = 95\]

From all these permutations, the maximum sum we get is 95. Therefore, the largest number that they could make is \(\boxed{95}\).  

\\ \bottomrule
\end{tabular}
\end{table}

\section{Limitations}
In this paper, we train a general reasoning VLM, namely \model, and test its effectiveness on a variety of benchmarks.
Despite it, here are few limitations about this work:

\begin{itemize}
    \item This paper focuses more on improving the visual reasoning capability of VLMs while neglecting the perception capability. More experiments on perception-related tasks and benchmarks should be an important future direction of this paper.
    As the proposed approach is general, the perception ability is easy to be enhanced by involving related training data, and feeding into our well-organized data processing pipeline.
    Besides, the improvement on reasoning ability may be also helpful to the perception ability. More evaluation experiments should be done in future work.

    \item Limited by the computational resource, the experiments are mostly conducted on Qwen2.5-7B-Instruct, without the test on larger and other VLMs. Since the proposed approach is general, it should be promising to improve the performance on other models with different architecture and scale. 

    \item More experiments on other advanced RL algorithms should be conducted, \eg DAPO~\citep{yu2025dapo}. In this paper, limited by computational resource, we only test the effectiveness on GRPO, the most popular RL algorithm. 
\end{itemize}

\end{document}